\documentclass[twocolumn]{article}
\usepackage{graphicx} 
\usepackage[round]{natbib}
\usepackage{url}
\usepackage{booktabs}

\title{Bridging the Usability Gap: Lessons from Interpreting Studies for Machine Interpreting Design}

\author{Claudio Fantinuoli \\
University of Mainz \\
fantinuoli@uni-mainz.de}
\date{}

\begin{document}

\maketitle
\section{Abstract}
Machine interpreting (MI), the live, real-time branch of speech translation, has achieved remarkable progress on standard benchmarks, with some systems approaching human parity on textual fidelity. Yet the user experience remains far inferior to interpreter-mediated communication, revealing what we term the \emph{accuracy illusion}: systems that appear accurate on paper but fail in practice to support smooth, goal-oriented interaction. This paper defines MI as a distinct subfield of speech translation, with its own characteristics and the need for evaluation methods grounded in communicative effectiveness rather than isolated fidelity metrics. Drawing on insights from interpreting studies, we identify critical dimensions of professional interpreting practice that are overlooked by current systems, and consolidate them into three interdependent design priorities for future MI: \emph{agency} (context-sensitive initiative and repair), \emph{grounding} (multimodal and discourse-level situational awareness), and \emph{experience} (adaptive improvement through real interaction). Together, these priorities chart a path toward closing the usability gap and enabling systems that can sustain authentic multilingual communication in real time.

\section{Introduction}

Speech translation (ST) has advanced rapidly in both research and deployment \citep{abdulmuminFindingsIWSLT20252025}, with systems reporting competitive scores on established metrics \citep{fantinuoliEvaluationAutomaticSimultaneous2021,weinBarriersEffectiveEvaluation2024} and some even framing their results in terms of human parity \citep{chengAchievingHumanParity2024a}. Yet users consistently report that live, interpreter-like performance remains far inferior to professional practice, often by a wide margin \citep[e.g.][]{korybskiExperimentingMachineInterpreting2026,matsushitaUnderstandingAIInterpreting2026}. This persistent discrepancy --- high benchmark accuracy versus low satisfaction in authentic use --- reveals a usability gap. We argue that the gap stems from what we call an \emph{accuracy illusion}: systems excel on standardized tests, where they are optimized for sentence-level fidelity, but fail to capture the interactional, contextual, and strategic dimensions that underpin successful interpreter-mediated communication.
 
For example, in a business meeting, an ST system may faithfully render figures and proper names, even outperforming a human reference on a post-hoc ``informativeness'' metric. In the actual interaction, however, the system lags two turns behind, fails to resolve a deictic expression such as ``this proposal'', and neutralizes rhetorical cues that signal whether a decision is tentative or final. As a result, participants interrupt, backtrack, and ultimately revert to lingua-franca English. While the translation may appear ``accurate'' on paper, in practice coordination breaks down. This discrepancy indicates that high-quality MI cannot be attained by optimizing for, or evaluating against, literal fidelity alone---yet lexical fidelity metrics remain the dominant optimization target in system development and evaluation campaigns \citep{fantinuoliEvaluationAutomaticSimultaneous2021,weinBarriersEffectiveEvaluation2024}.

We identify two reinforcing sources of misalignment that together produce this usability gap, both rooted in assumptions inherited from general-purpose ST research that do not transfer well to the specific demands of live interpreter-mediated communication:
(i) \emph{Reductive evaluation}: community practice privileges isolated linguistic accuracy (BLEU-like MT scores, WER, and their derivatives), which are weak proxies for communicative success in live interaction \citep{fantinuoliWang2024}---and where correlations with human judgments do exist, they concern textual accuracy rather than interactional adequacy \citep{machacekMTMetrics2023}; (ii) \emph{Problem framing}: engineering efforts often treat MI as a standard NLP mapping problem, rather than as an interactional task that must manage latency, initiative, repair, and audience design \citep[cf.][]{savoldiTranslationHandsMany2025}.

We argue that \emph{machine interpreting} (MI) --- the live, speech-to-speech application designed to support communication between speakers in real events --- requires a problem framing distinct from generic ST. Success in MI is not primarily measured by maximizing textual correspondence but by sustaining smooth, goal-directed interaction under immediacy constraints. This involves sensitivity to participants, roles, stakes, turn-taking, and multimodal context, as well as robust strategies for managing uncertainty and error.

This paper makes three main contributions:

\begin{enumerate}
    \item We propose a \textbf{definition of MI} as a distinct subfield of speech-to-speech translation, characterized by \emph{immediacy} and \emph{interactional embeddedness} (consecutive or simultaneous output intended for immediate use within live communicative events, without the possibility of post-editing or revision) and evaluated by communicative effectiveness rather than textual similarity.
    \item We \textbf{identify five critical features of human interpreting} that are under-modeled in current systems and whose absence hinders successful communication: faithfulness as meaning preservation (beyond literalism), fluency under real-time conditions, operational flexibility, situational awareness, cultural adaptation, and error management. These features are selected not as an exhaustive taxonomy but as a representative set that collectively traces the gap between accurate output and genuine communicative adequacy.
    \item We introduce a high-level \textbf{human-inspired design triad} for advancing MI: \emph{agency} (context-sensitive initiative and repair), \emph{grounding} (discourse and multimodal alignment of utterances), and \emph{experience} (learning from interaction over time). We make explicit how the five features map onto these three priorities and outline pathways for operationalizing them with current AI toolchains.
\end{enumerate}

Our aim is not to dismiss commonly used accuracy metrics but to situate them within a broader notion of success: whether interlocutors understand one another, coordinate action, and achieve intended outcomes with minimal disruption. This orientation builds on interpreting studies' accounts of professional practice \citep[e.g.][]{roderickInterpretingCommunicationProfession2014,pochhackerIntroducingInterpretingStudies2016,wadensjoInterpretingInteraction2014} and recent calls from the ST community to align system goals with interactional realities \citep{sperberMachineInterpretingLessons2025}.

The remainder of the paper is structured as follows: Section~\ref{sec:terminology} defines MI and differentiates it from other S2ST applications. Section~\ref{sec:overlooked} surfaces five overlooked features of human interpreting that matter for the usability of MI systems. Building on these, Section~\ref{sec:priorities} articulates three design priorities---agency, grounding, and learning from experience---and discusses how they can be implemented within contemporary pipelines. We conclude with implications for evaluation, system architecture, and interdisciplinary collaboration.

\section{Framing Machine Interpreting}\label{sec:terminology}
 
Before outlining the specific requirements of machine interpreting, it is helpful to situate the term within the broader landscape of speech-to-speech translation. The notion of machine interpreting is often used loosely, yet its scope and defining features remain contested. Establishing a working definition therefore serves not only to distinguish it from adjacent technologies, but also to frame the subsequent discussion of its aims, constraints, and implications.
 
\begin{quote}
\textbf{Machine interpreting} refers to a form of speech-to-speech translation whose defining characteristics are \emph{immediacy} and \emph{interactional embeddedness}: the translated output is produced in real time---whether in consecutive or simultaneous mode---within dynamic communicative settings where a human interpreter would traditionally be required, intended for immediate use and without the possibility of post-editing or revision.\footnote{For the purposes of this paper, we limit our definition to spoken languages, while recognizing that MI can also apply to signed languages.}
\end{quote}
  
The first defining characteristic, \emph{immediacy}, captures the temporal constraint under which MI operates: output must be produced in real time, without the possibility of post-editing or revision \citep{papiHowRealYour2025}. This sets MI apart from offline translation tasks and places it in direct continuity with human interpreting, where the pace of production is determined by the unfolding communicative event rather than by the translator. These two modes impose distinct constraints on system design:
 
\begin{itemize}
    \item \textit{Consecutive}: the system processes and translates entire segments (e.g., sentences or turns) after they are spoken.
    \item \textit{Simultaneous}: translation is generated incrementally while the source speech is still unfolding, requiring the system to act under partial information and to balance latency, accuracy, and fluency.
\end{itemize}
  
The second defining characteristic, \emph{interactional embeddedness}, concerns the communicative context in which MI operates rather than its temporal mode.\footnote{The notion of \emph{interactional embeddedness} builds on but refines \citet{pochhackerMachineInterpretingInterpreting2024}'s concept of \emph{situatedness}, which he pairs with embodiment as a defining condition of interpreting. We retain the situational dimension---the requirement that MI operate within and respond to a live communicative event---while foregrounding its interactional character rather than its physical or cognitive-embodied aspects. This distinction is consequential: as we argue in Section~\ref{sec:priorities}, effective MI requires active participation in interaction, but not necessarily a physical body.} While immediacy is a necessary condition for MI, it is not sufficient to distinguish it from the broader class of speech-to-speech translation (S2ST) systems. S2ST is a broad technical label for systems that convert spoken input into spoken output across languages, encompassing tasks such as offline dubbing and voice-over \citep{seligman_interactive_1997,papi_direct_2023,hu_neural_2022}. While all MI can be described as S2ST, not all S2ST qualifies as MI. The distinguishing criterion is interactional embeddedness: MI is designed specifically for live communicative situations---business negotiations, medical consultations, conference proceedings, community settings---in which the translation must sustain turn-taking, manage participant roles, and contribute to a shared communicative outcome. Offline dubbing or asynchronous voice-over tasks do not impose these demands.
 
Together, immediacy and interactional embeddedness give MI distinctive success criteria: not only accuracy, but also timing, interactional coherence, and contextual appropriateness. They also motivate two design principles that, while not part of the definitional core, are indispensable for any system that must function as an active agent in live communication: \emph{multimodality} and \emph{agency}. We develop both in Section~\ref{sec:priorities}.
  
Because MI is designed for live communicative situations traditionally handled by interpreters, benchmarks based solely on sentence-level accuracy are inadequate. Interpreters offer an inspirational reference point: they show that communicative success is not achieved through word-for-word fidelity, but when participants understand one another, coordinate action, and accomplish intended goals with minimal disruption. Such effectiveness relies on pragmatic adequacy, situational awareness, and interactional coherence---dimensions largely absent from current evaluation practices. The limited dialogue between computational research and Interpreting Studies \citep{pochhackerMachineInterpretingInterpreting2024} has reinforced the assumption that higher scores on BLEU or WER automatically imply communicative success, thereby contributing to the \emph{usability gap} and \emph{accuracy illusion} discussed earlier.
 
In sum, framing MI as a distinct form of speech-to-speech translation — defined by immediacy and interactional embeddedness, and evaluated by communicative effectiveness rather than textual similarity — situates it within broader disciplinary debates and underscores the need for criteria that move beyond sentence-level accuracy. In the following sections, we build on this foundation to examine what professional interpreting practice can teach us about the dimensions that current MI systems lack, and how those insights can guide the design of more communicatively adequate systems.

\section{Undermodeled Features of Human Interpreting}\label{sec:overlooked}

Despite rapid advances in speech translation (ST), most research and development has concentrated on linguistic transfer and benchmark optimization. This focus has produced impressive evaluation scores, yet it leaves largely unaddressed the broader set of capabilities that make interpreter-mediated communication effective in practice. Interpreting Studies provides a substantial body of knowledge on these capabilities, but its relevance to MI has thus far been only marginally integrated into system design and assessment \citep{sperberMachineInterpretingLessons2025, pochhackerMachineInterpretingInterpreting2024}.

In this section, we highlight five features of human interpreting that remain insufficiently modeled in current MI systems. These five are not intended as an exhaustive taxonomy of interpreting competence; rather, they are selected because each corresponds to a well-documented dimension of professional practice that is both consequential for communicative success and largely absent from current system design. Taken together, they trace a path from accurate output toward genuine communicative adequacy, and they provide the empirical grounding for the three design priorities developed in Section~\ref{sec:priorities}. As we argue there, the five features do not map one-to-one onto the three priorities: \emph{agency} consolidates faithfulness, fluency, operational flexibility, cultural adaptation, and error management---all of which are expressions of the system's capacity for context-sensitive, goal-directed action; \emph{grounding} operationalizes situational awareness through multimodal and discourse-level perception; and \emph{experience} is a second-order priority that enables both agency and grounding to improve over time.

\subsection{Faithfulness}

\textit{Faithfulness} in interpreting does not mean literal repetition, but rather the preservation of the speaker's intended meaning---even if this requires reformulation, omission, or correction \citep{gileBasicConceptsModels2009,jonesConferenceInterpretingExplained2014,pochhackerIntroducingInterpretingStudies2016}. This conception of faithfulness differs fundamentally from literal translation approaches \citep[cf.][]{newmarkApproachesTranslation1981}: what matters is not surface correspondence but pragmatic equivalence. As \citet{jonesConferenceInterpretingExplained2014} puts it, the interpreter's goal is to say what the source speaker \emph{would have said} in the target language, had they been fluent in it---a formulation that captures the productive tension at the heart of professional practice, where fidelity to the speaker's intent may require departing from their words. Professional interpreters routinely adapt cultural references, silently correct slips of the tongue, or condense detail when full reproduction would risk overloading listeners, all in service of preserving the speaker's communicative purpose rather than their literal output.

\textbf{Implications for MI:} Current systems tend to equate quality with lexical fidelity, but for real-world usability MI must be optimized for pragmatic equivalence. This means generating output that preserves communicative intent under live conditions---balancing precision with clarity---and integrating evaluation metrics that capture meaning preservation rather than only word-level accuracy. It also implies that deviations from the source text are not necessarily errors: an MI system capable of distinguishing between a speaker's words and their intent, and of acting on that distinction, would represent a significant advance over current architectures.

\subsection{Fluency}

\textit{Fluency} ensures that output is not only grammatically correct but also coherent, natural, and easy to follow. This involves segmenting complex sentences, handling discourse markers, and maintaining prosody and intonation so that content remains accessible in real time \citep{gillesConsecutiveInterpretingShort2019,pochhackerIntroducingInterpretingStudies2016}. Research on interpreting quality has consistently identified fluency as one of the most important listener-oriented criteria: it ranks among the top priorities for both conference interpreters and end users, alongside accuracy and completeness \citep{buhlerLinguisticCriteria1986,pochhackerIntroducingInterpretingStudies2016}. Crucially, fluency in interpreting is not reducible to the absence of disfluencies; it encompasses the prosodic organization of output---tempo, pause placement, intonation---which directly affects listener comprehension and perceived accuracy, even independently of the content being conveyed \citep{shlesingerProsodicDimensionsSI1994}.

\textbf{Implications for MI:} Current MI systems treat fluency as a byproduct of accurate transcription and translation, with no explicit modeling of the prosodic and temporal dimensions that determine how output is experienced by listeners. Yet fluency in live interpreting is not a surface property of text but an interactional achievement---produced for a specific audience, under time pressure, with deliberate control of pace, pause, and intonation to manage listener comprehension. A system that generates grammatically correct output but delivers it with flat prosody, poorly placed pauses, or misaligned tempo will impede communication even when its content is accurate. Incorporating explicit fluency modeling---including prosodic generation, delivery pacing, and audience-sensitive output shaping---is therefore a missing design dimension in current MI architectures.

\subsection{Operational Flexibility}

Human interpreters adapt dynamically to event type, speaker style, pace, and communicative goals. They may complete sentences when speakers break off, slow down or simplify when addressing non-experts, or shift strategies when interpreting written text read aloud \citep{wadensjoInterpretingInteraction2014}. In healthcare or community settings, they may further assume different roles, such as system agent, cultural broker, or integration agent---depending on situational needs \citep{leanzaWorkingInterpreters2014}.  

\textbf{Implications for MI:} Current MI systems are architecturally rigid: they operate with fixed pipelines optimised for a single mode of delivery, with no capacity to adjust strategy in response to speaker pace, domain, audience expertise, or event type. This rigidity is a fundamental design limitation, not a performance gap. A system that cannot modulate its behaviour when a speaker accelerates, switches register, or reads from a prepared text will degrade in precisely the heterogeneous conditions that characterise real-world deployment. Operational flexibility must therefore be built into MI architectures as a first-class design requirement---not treated as a refinement to be added once core accuracy is achieved.

\subsection{Situational Awareness}

Interpreters operate with a heightened awareness of context: they attend to visual cues, facial expressions, gestures, intonation, and participant relationships. Beyond processing words, they perceive the emotional tenor, interactional goals, and the broader environment in which communication unfolds \citep{endsleySituationAwarenessHuman1999,viaggioQuestOptimalRelevance2002}. This situational awareness informs decisions about register, nuance, and when (or whether) to intervene.  

\textbf{Implications for MI:} Current MI systems are perceptually impoverished: they process the acoustic signal in isolation, with no access to the visual, prosodic, and relational cues that human interpreters rely on to disambiguate meaning and calibrate output. This is not merely a missing feature but a structural constraint that renders current systems blind to entire dimensions of communicative context. The result is output that may be locally accurate but globally incoherent---failing to resolve referents, misreading the interactional stakes, or producing register-inappropriate translations because the system has no model of the situation in which it is operating. Situational awareness must therefore be understood as a core architectural requirement for MI, not an optional enrichment.

\subsection{Cultural Adaptation}

Interpreters routinely mediate cultural differences, ensuring not only that words are translated but also that intent is preserved. They substitute idioms, explain references, and adjust pragmatic force depending on audience and context \citep{pochhackerIntroducingInterpretingStudies2016,ramirezReflectiveAssignmentsHealth2025}. For instance, the Italian phrase ``In bocca al lupo'' is rendered as ``Good luck'' in English rather than through a literal reference to wolves.  

\textbf{Implications for MI:} Current MI systems lack any principled mechanism for cultural mediation. They may incidentally handle high-frequency idioms through exposure in training data, but this is pattern matching, not adaptation: the system has no model of the cultural distance between source and target, no awareness of the audience's background, and no capacity to decide when literal rendering, functional equivalence, or explicit explanation is called for. The consequence is output that can mislead or alienate users in precisely the cross-cultural settings where MI is most needed. Cultural adaptation must therefore be treated as an active design target---requiring explicit knowledge representations, audience modeling, and context-sensitive decision policies---rather than a side effect of scale.

\subsection{Error Management}

Even expert interpreters encounter noise, ambiguities, or cognitive overload, to name just a few. What distinguishes professional practice is the use of robust recovery strategies: reformulation, simplification, omission, or, when necessary, explicit repair requests \citep{gileBasicConceptsModels2009,jonesConferenceInterpretingExplained2014}. Interpreters can smooth over hesitations, improvise under pressure, or diplomatically moderate hostile remarks while still preserving communicative intent.  

\textbf{Implications for MI:} Current MI systems have no principled error management layer. When faced with low-confidence input, they typically default to the most probable output regardless of communicative risk---producing fluent-sounding but potentially misleading translations with no signal to the listener that uncertainty is present. Unlike human interpreters, who can signal doubt, request clarification, simplify under pressure, or explicitly flag a mishearing, current systems have no repertoire of recovery strategies and no mechanism for deciding when to deploy them. This is a critical gap: in high-stakes domains such as medical consultations or legal proceedings, the absence of error management is not merely a quality issue but a safety one. Building explicit uncertainty handling and repair strategies into MI architectures is therefore not a refinement but a prerequisite for responsible deployment.

\paragraph{}
The five features examined above---faithfulness as pragmatic equivalence, fluency as interactional achievement, operational flexibility, situational awareness, and cultural adaptation and error management---identify specific dimensions of human interpreting performance that are systematically absent from current MI systems. They are not peripheral refinements: each one corresponds to a capability without which interpreter-mediated communication would fail in practice, and each one is either unmodeled or only superficially addressed in existing architectures. Taken together, they make visible the distance between what current MI systems do and what effective live multilingual communication requires. This gap motivates three overarching design goals---\emph{agency}, \emph{grounding}, and \emph{experience}---which we develop in the following section as an actionable framework for advancing MI beyond benchmark optimisation toward genuine communicative adequacy.

\section{Human-Inspired Design Priorities for Machine Interpreting}\label{sec:priorities}

This section develops three interdependent design priorities for MI---\emph{agency}, \emph{grounding}, and \emph{experience}---each grounded in the interpreting features identified above. They are mutually reinforcing: agency relies on contextual grounding, grounding becomes meaningful when enacted through agentic choices, and both are strengthened when informed by accumulated experience.

The mapping from five features to three priorities reflects a deliberate consolidation. \emph{Agency} subsumes faithfulness (as meaning preservation rather than literal transfer), fluency (as listener-oriented output management), operational flexibility (as adaptive decision-making), cultural adaptation (as pragmatic mediation), and error management (as repair and uncertainty handling): all five are expressions of the system's capacity to take context-sensitive, goal-directed action during an interaction. \emph{Grounding} operationalizes situational awareness by equipping the system with the multimodal and discourse-level perception that makes agency reliable. \emph{Experience} is a second-order priority: it enables both agency and grounding to improve over time through exposure to diverse communicative events, rather than remaining static.

\subsection{Agency}\label{subsec:agency}

Agency refers to the system's ability to take proactive, context-sensitive decisions during an interaction. We treat agency as a functional property---shaping communication so that it remains effective under immediacy constraints---rather than as a claim about human-like intentionality or consciousness \citep{fantinuoliInterpretingWithoutIntelligence2025}. This functional perspective aligns with how professional interpreters exercise agency in practice: they initiate repairs, manage turn-taking, and adapt register to audience expectations in order to sustain effective interaction \citep{wadensjoInterpretingInteraction2014,pochhackerIntroducingInterpretingStudies2016}. In the context of MI, the challenge is to approximate such context-sensitive choices within contemporary AI architectures.

Formally, agency can be defined through three fundamental criteria: \emph{interactivity}, \emph{autonomy}, and \emph{adaptability}. This tripartite definition is widely recognized in the literature on artificial agents \citep{franklinItAgentJust1997,wooldridgeIntelligentAgentsTheory1995}. \textit{Interactivity} refers to an agent's capacity to engage with its environment through mutual influence, acting upon it and, in turn, being acted upon. \textit{Autonomy} denotes the ability to initiate state changes without direct external causation---not complete independence from the environment, but the capacity for self-initiated and directed action within given constraints \citep{russellArtificialIntelligenceModern2022}. \textit{Adaptability} is the capacity to adjust behaviour in response to input, such as changing conditions or newly acquired information \citep{jenningsRoadmapAgentResearch}. We treat this tripartite framework as a design target rather than a description of existing systems: current MI architectures exhibit these properties only partially, and approximating them more fully is precisely what motivates the directions outlined below.

In MI systems, agency can manifest in multiple ways. Current ST architectures typically handle uncertainty by selecting the most probable continuation. By contrast, an agentic MI system would actively monitor for signs of communicative trouble (e.g., noisy input, discourse incoherence), weigh the trade-off between error and delay, and decide whether to proceed, hedge, omit, or seek clarification. Beyond managing uncertainty, agency also entails pragmatic and cultural adaptation: rendering idioms in target-language equivalents or explaining them, explicating culture-bound references, or adjusting register to match the communicative setting, whether a clinic, a classroom, or a boardroom. 

Below we organize typical agentic behaviors as observable \emph{interactional} choices rather than mere fixed text transformations:
\begin{itemize}
    \item \textbf{Uncertainty management:} mark speaker hedges; propagate calibrated confidence; avoid unwarranted factual commitments when evidence is weak.
    \item \textbf{Repair initiation:} request repetition/rephrasing; signal mishearing; ask for definitions of terms-of-art when stakes are high.
    \item \textbf{Cultural-pragmatic mediation:} explicate culture-specific items; preserve rhetorical force (humour, wordplay) with target-appropriate devices where feasible; avoid over-adaptation in diplomatic/strategic talk.
    \item \textbf{Register and audience control:} match formality and technicality to roles and goals (clinician--patient vs.\ CEO--board); simplify or scaffold explanations for non-experts.
    \item \textbf{Ethically constrained intervention:} avoid sanitizing or intensifying sensitive language unless mandated by context; surface potential harms; prefer least-invasive choices in high-stakes domains.
    \item \textbf{Revision-on-feedback:} incorporate listener/speaker feedback to amend earlier output and repair downstream misunderstandings.
\end{itemize}

Such behaviours rest on several core design ingredients. We focus on three as an exemplification: (i) continuous monitoring of ASR and translation confidence, (ii) explicit decision policies for negotiating the trade-off between accuracy and latency (for simultaneous modality), and (iii) pragmatic guardrails anchored in domain-specific communicative norms. Understood this way, agency transforms MI from a passive conduit of speech-to-speech transfer into an adaptive participant capable of sustaining interaction under real-world conditions.

\textbf{Design Directions:} To operationalize agency, current approaches can be extended to  
(a) integrate real-time monitoring modules for uncertainty, including ASR confidence estimation \citep{jiangConfidenceScoringAutomatic2005}, quality estimation for MT \citep{speciaQualityEstimationMachine2018}, and end-of-turn detection for interactional management \citep{skantzeTurntakingConversationalSystems2021};  
(b) implement adaptive decision policies that balance accuracy and latency, drawing on research in simultaneous translation strategies such as reinforcement learning for read/write scheduling \citep{grissomiiDontFinalVerb2014}, adaptive delay policies \citep{guLearningTranslateRealtime2017a}, prefix-to-prefix anticipation \citep{maSTACL2020}, and divergence-based adaptive scheduling \citep{arivazhaganMILk2019,zhaoAdaptivePolicy2023,sunAdaptiveSignSST2024};  
(c) embed pragmatic guardrails, including politeness and formality control \citep{niuPoliteDialogueGeneration2018,raoControlFormalityNeural2018}, idiom and proverb adaptation \citep{zouNeuralIdiomaticity2021,rizzoCulturalAdaptationNMT2023}, and culturally sensitive paraphrasing strategies; and  
(d) exploit large language models not only as translation engines but also as controllable agents, where prompting and system instructions steer hedging, explicitation, register adaptation, and user-sensitive reformulation without retraining \citep{plaataAgenticLLMSurvey2024,nisaAgenticAI2025}.\footnote{To illustrate how such control might be operationalized, consider a system instruction of the following form: \textit{``You are assisting live speech-to-speech interpreting. Maintain communicative effectiveness under latency constraints. If ASR\_confidence $<0.75$ or MT\_QE $<0.6$, prefer conservative paraphrase and add a brief hedge (e.g., `approximately'), or omit low-confidence spans. If end-of-turn is uncertain, insert a micro-pause ($\leq$ 400\,ms) before committing to disambiguating content. If domain = clinical and audience = lay, downshift register and explicate acronyms on first mention.''} This is a proof-of-concept illustration; specific thresholds and policies would require empirical calibration for each deployment context.}

Taken together, these directions point toward MI systems that behave less like rigid pipelines and more like adaptive agents, capable of responding to uncertainty, interactional contingencies, and cultural expectations.

\subsection{Grounding}\label{subsec:grounding}

Grounding means situating translation in the unfolding communicative event. Human interpreters do not rely on words alone; they sustain \emph{common ground} by tracking discourse, interpreting gestures, and anticipating participants' goals \citep{wadensjoInterpretingInteraction2014,setton_simultaneous_1999}.  

Two dimensions are especially relevant for MI. First, \emph{discourse grounding}: keeping track of dialogue history to resolve references (``this proposal''), manage ellipsis, or adapt register. Second, \emph{multimodal grounding}: using situational cues such as gaze, slides, or prosody to disambiguate meaning. Without these capacities, output risks being formally correct but pragmatically incoherent.  

Technically, LLMs with memory augmentation offer discourse sensitivity, while multimodal AI can provide contextual metadata (e.g., who is speaking, what is being pointed at). Although vision systems remain imperfect, they are already sufficient to improve referent resolution and adaptation. Grounding thus provides the connective tissue that makes agency reliable and user-oriented.

\textbf{Design Directions:} Progress on grounding could involve  
(a) equipping MI with dialogue-state tracking to maintain context across turns \citep{williamsDSTC2016},  
(b) integrating multimodal input channels, building on visual grounding in translation \citep{elliottFindingsSharedTask2017,caglayanMultimodalMT2019} and slide/figure understanding \citep{masryChartQA2022,tanSlideVQA2023},  
and (c) exploiting paralinguistic and prosodic features for contextualization \citep{eybenopensmile2010}.  

\subsection{Experience}\label{subsec:experience}

Finally, MI systems need to learn from experience rather than remain static. Human interpreters acquire expertise through repeated exposure to diverse communicative events, developing repertoires of strategies and the flexibility to adapt to the unexpected \citep{moser-mercer_remote_2005-1,gileBasicConceptsModels2009}.

This matters because communicative meaning is not a fixed property of linguistic forms. As Wittgenstein argued, meaning arises from use in specific contexts, not from words alone \citep{wittgensteinPhilosophicalInvestigationsPhilosophische1989}. The same phrase may signal different intentions depending on culture, tone, or situation. Human interpreters cope with this variability by drawing on accumulated interactional experience---recognizing recurring patterns, anticipating domain-specific conventions, and adjusting strategies accordingly. An MI system that remains static after deployment cannot replicate this capacity: it will encounter speaker styles, domain registers, and interactional configurations that its training distribution did not anticipate. The experience priority addresses precisely this brittleness. It transforms the system from a fixed artifact into an adaptive one that improves through exposure---meaning that agency and grounding, however well-designed at deployment, must be revisable in light of feedback from real interactions. This is what distinguishes an MI system that merely performs from one that develops.

Reinforcement learning offers one pathway: systems can refine segmentation or latency strategies based on feedback \citep{grissomiiDontFinalVerb2014,guLearningTranslateRealtime2017a,sperber_speech_2020}. More ambitiously, MI could build an \emph{interactional memory} that recognizes recurring patterns and adapts accordingly. Without such mechanisms, MI risks remaining brittle---fluent only under ideal conditions. With them, systems could begin to approximate the adaptability that characterizes human interpreters.

\textbf{Design Directions:} To enable experience-driven improvement, MI systems could  
(a) integrate RL-based policies for latency and segmentation \citep{guLearningTranslateRealtime2017a,maSTACL2020},  
(b) apply continual and parameter-efficient learning to prevent catastrophic forgetting \citep{kirkpatrickOvercomingCatastrophicForgetting2017,huLoRA2021}, and  
(c) leverage human-in-the-loop feedback via interactive MT frameworks \citep{greenTranslatingGeneralPurposE2014,kreutzerBanditStructuredPrediction2018}.  

\section{Discussion}

The analysis in the preceding sections yields two sets of implications---one for evaluation and one for system design and disciplinary dialogue---that we address in turn.

\subsection*{Implications for Evaluation}

A central argument of this paper is that the metrics currently dominant in the ST and MI literature---BLEU, WER, and their derivatives---are insufficient proxies for communicative success in live interaction. This is not merely a technical limitation but a conceptual one: these metrics are designed to measure correspondence between a system output and a reference translation, but they presuppose that such a reference adequately captures communicative quality. In the context of MI, where success is measured by whether interlocutors understand one another and achieve their goals, this presupposition fails. A translation may score highly on BLEU while failing to resolve a deictic reference, neutralizing a rhetorical hedge, or delivering output too slowly for real-time coordination.

Progress therefore requires evaluation frameworks anchored in communicative effectiveness. Several directions are worth pursuing. Task-based evaluation---measuring whether participants achieve communicative goals in controlled interactions---provides the most direct index of MI quality but is costly and difficult to standardize \citep{weinBarriersEffectiveEvaluation2024}. Complementarily, dimension-specific metrics addressing fluency under time pressure, repair frequency, reference resolution accuracy, and register appropriateness could provide actionable diagnostic feedback at scale. Evaluation should also be sensitive to mode: consecutive and simultaneous MI impose different constraints and should not be assessed against a single benchmark. Crucially, any evaluation framework for MI must incorporate the interactional perspective developed in this paper---the question is not only whether individual utterances are accurate, but whether the system sustains coherent, goal-directed communication across turns.

\subsection*{Implications for System Design and Disciplinary Dialogue}

Framing MI as interactionally embedded speech-to-speech translation---rather than as a special case of offline ST---redirects design attention toward the dimensions that matter for live use: latency management, discourse tracking, multimodal input, and adaptive decision-making. The agency--grounding--experience triad proposed here is intended as a practical scaffold for this reorientation.

Framed in this way, our definition enters into dialogue with ongoing debates in Interpreting Studies. \citet{pochhackerMachineInterpretingInterpreting2024}, for example, questions whether MI qualifies as ``interpreting'' at all, arguing that \emph{embodiment} is a necessary condition. We build on this concern but propose a broader notion of \emph{interactional embeddedness}, conceptualized through \textit{multimodality} and \textit{agency}. Effective interpreting---whether human or machine---depends on the ability to mobilize communicative resources beyond the lexical signal, yet it does not strictly require a physical body. What is indispensable is a multimodal presence capable of sustaining interaction; embodiment in the literal sense is not.

Realizing this vision will require sustained dialogue between Interpreting Studies and computational research---two communities that have historically operated in parallel \citep{pochhackerMachineInterpretingInterpreting2024,sperberMachineInterpretingLessons2025}. The five features identified in Section~\ref{sec:overlooked} are drawn from established IS research; translating them into the three computational design priorities of Section~\ref{sec:priorities} is precisely the kind of interdisciplinary work that can move MI beyond benchmark optimization toward genuine communicative adequacy.

\section{Conclusions}\label{conclusions}
 
In this paper, we have argued that the usability gap between advances in speech translation technology and its successful deployment in real-world settings stems from overlooked factors that are crucial to interpreter-mediated communication. We contend that the most promising path forward for Machine Interpreting (MI) is to examine what makes human interpreters the gold standard of live multilingual communication.
 
Human interpreters are not mere linguistic conduits but cognitive and social architects of interaction. They grasp not only what is said but also why it is said, how it is intended to be received, and by whom. Their work spans cognitive skills (e.g., memory, anticipation, analysis), social competences (e.g., mediating relationships, bridging cultural gaps), and ethical responsibilities (e.g., impartiality, advocacy). This multifaceted role enables them to shape communication toward mutual understanding and desired outcomes---capacities that current speech translation systems do not yet achieve.
 
Future MI systems will therefore require a paradigm shift: from narrow language processing to a holistic engagement with the communicative event, incorporating elements of cognitive modeling, social intelligence, and cultural awareness. To approximate this level of success, we proposed focusing research on three directions. \emph{Agency} consolidates five key interpreting features---faithfulness, fluency, operational flexibility, cultural adaptation, and error management---into a unified capacity for context-sensitive initiative. \emph{Grounding} operationalizes situational awareness through multimodal and discourse-level processing. \emph{Experience} enables both to improve over time through exposure to real interactions, transforming a static system into one that develops. Together, these three priorities define a human-inspired roadmap for MI that addresses the specific shortcomings of current systems while remaining tractable with contemporary AI toolchains.
 
A caveat is worth stating explicitly. The agency--grounding--experience triad is derived from what makes human interpreters successful, and it describes the full set of capabilities that a system would need to match human performance in live multilingual communication. Whether all three are necessary for MI to be \emph{useful} in practice is a separate and equally important question. It is plausible---and empirically testable---that even partial implementation of these priorities, or prioritization of one over the others depending on domain and use case, could yield significant improvements in real-world deployability. Medical consultations may demand stronger grounding and error management than large conference settings; community interpreting may require cultural adaptation more urgently than fluency modeling. The triad should therefore be read not as an all-or-nothing prerequisite but as a diagnostic framework: it identifies where current systems fall short and where targeted investment is most likely to close the usability gap in specific deployment contexts. Future empirical work will be needed to establish which combinations, and at what level of implementation, are sufficient for MI to be trusted and used effectively by real users in real situations.
\bibliography{paperMI}

\end{document}